\documentclass[sigconf]{acmart}

\usepackage{hyperref}
\usepackage{multirow}
\usepackage{tabularx}
\usepackage{url}
\usepackage{xspace}

\usepackage{subcaption}
\usepackage{graphicx}
\usepackage{tcolorbox} 
\usepackage{courier}
\usepackage{svg}
\usepackage{booktabs}
\usepackage{multirow}
\usepackage{enumitem}
\usepackage{ragged2e}
\AtBeginDocument{%
  }

\setcopyright{acmlicensed}
\copyrightyear{2018}
\acmYear{2018}
\acmDOI{XXXXXXX.XXXXXXX}
\acmConference[Conference acronym 'XX]{Make sure to enter the correct
  conference title from your rights confirmation email}{June 03--05,
  2018}{Woodstock, NY}
\acmISBN{978-1-4503-XXXX-X/2018/06}

\begin{document}

\title{REGEN: A Dataset and Benchmarks with Natural Language Critiques and Narratives}

\author{Kun Su}
\affiliation{%
    \institution{Google Research}
    \city{Mountain View}
    \state{CA}
    \country{USA}
}
\authornote{Equal contribution.}

\author{Krishna Sayana}
\authornotemark[1]
\affiliation{%
    \institution{Google Research}
    \city{Mountain View}
    \state{CA}
    \country{USA}
}

\author{Hubert Pham}
\authornotemark[1]
\affiliation{%
    \institution{Google Research}
    \city{Mountain View}
    \state{CA}
    \country{USA}
}

\author{James Pine}
\authornotemark[1]
\affiliation{%
    \institution{Google Research}
    \city{Mountain View}
    \state{CA}
    \country{USA}
}

\author{Yuri Vasilevski}
\authornotemark[1]
\affiliation{%
    \institution{Google Research}
    \city{Mountain View}
    \state{CA}
    \country{USA}
}

\author{Raghavendra Vasudeva}
\affiliation{%
    \institution{Google Research}
    \city{Mountain View}
    \state{CA}
    \country{USA}
}

\author{Marialena Kyriakidi}
\affiliation{%
    \institution{Google Research}
    \city{Mountain View}
    \state{CA}
    \country{USA}
}

\author{Liam Hebert}
\affiliation{%
    \institution{University of Waterloo}
    \city{Waterloo}
    \state{Ontario}
    \country{Canada}
}
\authornote{Work done while at Google Research.}

\author{Ambarish Jash}
\affiliation{%
    \institution{Google Research}
    \city{Mountain View}
    \state{CA}
    \country{USA}
}

\author{Anushya Subbiah}
\affiliation{%
    \institution{Google Research}
    \city{Mountain View}
    \state{CA}
    \country{USA}
}

\author{Sukhdeep Sodhi}
\affiliation{%
    \institution{Google Research}
    \city{Mountain View}
    \state{CA}
    \country{USA}
}

\renewcommand{\shortauthors}{Kun Su, Krishna Sayana, Hubert Pham, James Pine, Yuri Vasilevski et al.}

\begin{abstract}
This paper introduces a novel dataset REGEN (\textbf{R}eviews \textbf{E}nhanced with \textbf{GE}nerative \textbf{N}arratives), designed to benchmark the conversational capabilities of recommender Large Language Models (LLMs), addressing the limitations of existing datasets that primarily focus on sequential item prediction. REGEN extends the Amazon Product Reviews dataset by inpainting two key natural language features: (1) user critiques, representing user "steering" queries that lead to the selection of a subsequent item, and (2) narratives, rich textual outputs associated with each recommended item taking into account prior context. 
The narratives include product endorsements, purchase explanations, and summaries of user preferences.

Further, we establish an end-to-end modeling benchmark for the task of conversational recommendation, where models are trained to generate both recommendations and corresponding narratives conditioned on user history (items and critiques). For this joint task, we introduce a modeling framework LUMEN (\textbf{L}LM-based \textbf{U}nified \textbf{M}ulti-task Mod\textbf{e}l with Critiques, Recommendations, and \textbf{N}arratives) which uses an LLM as a backbone for critiquing, retrieval and generation. We also evaluate the dataset's quality using standard auto-rating techniques and benchmark it by training both traditional and LLM-based recommender models. Our results demonstrate that incorporating critiques enhances recommendation quality by enabling the recommender to learn language understanding and integrate it with recommendation signals. Furthermore, LLMs trained on our dataset effectively generate both recommendations and contextual narratives, achieving performance comparable to state-of-the-art recommenders and language models. 
\end{abstract}



\maketitle

\begin{figure*}[!ht]
    \centering
    \includegraphics[width=0.95\linewidth]{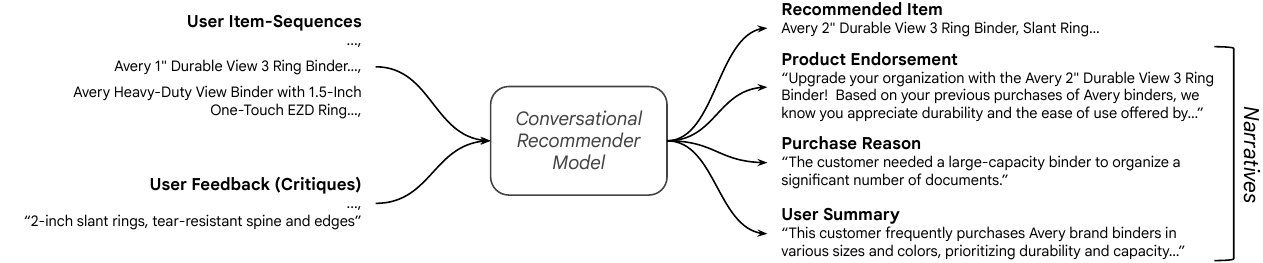}
    \caption{REGEN is a dataset for training conversational recommender models that output both item recommendations and engaging natural language, given user history and feedback. Based on the user item-sequences in Amazon Product Reviews, REGEN adds (1) inferred user feedback, in the form of natural language critiques, and (2) rich text narratives, in the form of product endorsements, purchase reasons, and user summaries---all personalized to each user.}
    \label{fig:regen_schematic}
\end{figure*}

\section{Introduction}

Applying Large Language Models (LLMs) to conversational recommenders~\cite{ning2024userllmefficientllmcontextualization,lc-rec,p5rec,zhu2023collaborative,liang-etal-2024-llm} has received some interest of late. In addition to providing personalized recommendations, conversational recommenders  engage in natural dialog with users e.g. to explain recommendations, refine recommendations based on natural language feedback from the user. LLMs show promise in fulfilling the necessary language capabilities of such recommenders.

This is challenging as existing datasets for training conversational recommender systems lack rich, contextual narratives and explicit user feedback. Such a dataset is crucial for modeling natural language interactions in the context of the recommended items, so generated text is consistent and relevant to the recommendations and user preferences. While several recent works explore modeling approaches that integrate the recommender's internal state with an LLM \cite{lc-rec, zhu2023collaborative, Hua_2023, userllm}, they are limited to existing datasets (e.g. \cite{p5rec, ni-etal, chen2024unlockingwhybuyingintroducing, wang2024bridging, li2017redial, liang-etal-2024-llm}) that concentrate on user item-sequences, short dialog snippets, structured outputs, or small summaries that lack the rich and varied conversational elements. 

Collecting such a dataset from users is challenging, and would typically require a large set of real users interacting with a conversational recommender. We thus continue the trend 
\cite{leszczynski2023talkwalksyntheticdata, ni-etal, pmlr-v162-dai22a}
of exploring the hypothesis that LLMs can (1) infer user preferences and goals from an existing dataset of user behavior traces; and (2) generate compelling and plausible user narratives and responses, consistent with ground truth behavior. Aside from expediency, an advantage to this approach is that we can augment existing popular sequential recommendation datasets.

We present REGEN (Reviews Enhanced with GEnerative Narratives), a dataset that augments the Amazon Reviews datasets \cite{ni-etal} by incorporating natural language outputs relevant to conversational recommendations (Figure~\ref{fig:regen_schematic}). REGEN adds diverse narratives such as purchase reasons, explanations, product endorsements, user summaries \& concise user profiles, personalized to each user. It also adds plausible critiques, or natural language feedback that the user issues in response to recommendations to steer the recommender. To augment the dataset, we use Gemini 1.5 Flash \cite{geminiteam2024gemini15unlockingmultimodal} with few shot prompting to inpaint narratives and critiques. The LLM generates those features using the user's rating history, product review text, and item descriptions in the original Amazon Reviews datasets. As an initial measure of feasibility, we automate quality inspection with an auto-rater LLM \cite{RoboRater}. The auto-rater assesses the generated outputs across multiple attributes, with a focus on grounding and factuality based on the user's historical interactions. 

In addition to a dataset, there is also a need for a single training task that integrates user behavior and recommendations, to evaluate a model's consistency and handling of those aspects. While many previous studies (e.g. \cite{userllm, p5rec})
propose models that are trained with separate individual objectives, e.g. to make recommendations, rating predictions, or review generation, we believe a task that combines those aspects is a critical step towards evaluating coherence in conversational recommendations. 
To aid model development, we propose a single-turn conversational task as a step towards fully conversational recommenders. In this task, the model input is a user history of items, combined with user critiques between some items. The output is the next item to recommend and rich natural language narratives consistent with the recommend item and user preferences. 
The task requires the model to effectively combine personalized sequential recommendation, critiquing, and natural language generation into a single, coherent turn of a fully conversational recommender.

Finally, to evaluate the utility of both the REGEN dataset and task, we present reference models and strong baselines. Here we introduce LUMEN (\textbf{L}LM-based \textbf{U}nified \textbf{M}ulti-task Mod\textbf{e}l with Critiques, Recommendations, and \textbf{N}arratives), a single model using an LLM backbone trained to to generate recommendations and narratives by integrating collaborative filtering signals, content signals, and natural-language critiques inputs. LUMEN leverages user-item interaction data, rich item content representations, and user feedback to generate outputs that are informative and aligned with user preferences. Our experiments establish benchmarks that demonstrate the effectiveness of this architecture in producing human-like conversational recommendations. 

In summary, the key contributions are:
\begin{itemize}[nosep]
    \item REGEN, an open-source dataset,\footnote{\url{https://www.kaggle.com/datasets/googleai/regen-reviews-enhanced-with-generative-narratives}} with critiques and rich natural language narratives that are consistent and personalized to a user's consumption history. We evaluate it for factuality, grounding, and accurately reflecting user context.
    \item A corresponding task towards a fully conversational recommender which can serve as a foundation for future research. 
    \item Reference models that show LLM architectures combining generative recommendation and language task achieve performance comparable to state-of-the-art recommenders and language models. 
    \item Validation of the model's ability to learn and generate contextually rich narratives and leverage user critiques to dynamically guide recommendations. We further demonstrate generation of coherent conversational traces.
\end{itemize}

\section{Related Work}
REGEN is inspired by existing datasets and modeling tasks
that support the development and evaluation of recommender systems.

\subsection{Datasets}
\subsubsection{Conversational}
Many conversational datasets are dialog-based, representing conversation between a user and an agent. For example, DuRecDial \cite{liu-etal-2020-towards-conversational, liu-etal-2021-durecdial} is a crowd-sourced, multi-domain dialogue dataset with pre-defined goals. E-ConvRec \cite{jia2022convrec} provides a large-scale conversational recommendation dataset based on real pre-sales dialogues between users and customer service. ReDial \cite{li2018conversational} is an annotated dataset of dialogues  of movie recommendations. TG-ReDial \cite{zhou2020topicguidedconversationalrecommender} extends ReDial by incorporating topic information to guide conversations and recommendations, while E-Redial \cite{10.1145/3539618.3591884} expands it with external knowledge to improve explainability. These datasets focus on constructing or collecting dialogues, but typically lack coupling between the dialogue and the user's underlying preferences or historical actions. LLM-Redial \cite{liang-etal-2024-llm} pursues a user-centric focus by creating dialogues via templates, rooted in real user profiles. REGEN is similar to LLM-Redial with its focus on maintaining consistency between the user profile and the narratives but differs in several ways. First, REGEN preserves the sequential item history to maintain temporal consistency, enabling it to generate user feedback and narratives conditioned on past user behavior. Second, rather than short dialogue turns, REGEN focuses on personalized, rich narratives that a knowledgeable conversational recommender might produce. We believe a narrative-focused dataset helps test a recommender's language understanding, in the context of the recommended items and user preferences.

\subsubsection{Narratives}
The ``Justifying Recommendations'' dataset by Ni et al. \cite{ni-etal} and the ``Unlocking the Why of Buying'' by Chen et al. \cite{chen2024unlockingwhybuyingintroducing} focus on generating explanations from reviews. The ``Amazon-C4'' dataset by Hou et al. \cite{hou2024bridginglanguageitemsretrieval} presents a new reviews dataset that constructs complex queries and associates them with items to evaluate the models on recommendation tasks. REGEN extends their work with open-ended narratives suited to several recommendation tasks, conditioned on the entire user history, and covering use cases beyond purchase explanations. REGEN targets natural language outputs that reflect the rapidly evolving capabilities of LLMs, with longer, more nuanced, and semi-structured and unstructured responses. 

\subsubsection{Critiquing}
Critiquing for classical recommender systems is surveyed in \citet{Chen2012-dv, Gao2021-xo, Jannach2021-bp}. Early critiquing systems enabled the user to provide feedback on recommended items, typically at the feature-level, through (e.g.) suggested pre-designed tweaks \citep{Burke1996-rv} or language-driven dialog systems \citep{Shimazu2002-uq, Thompson2004-jj, Grasch2013-kw}. 
Recent work \citep{Wu2019-qx, Luo2020-uw, Luo2020-qh} explores systems that jointly learn latent representations for users, items, and key-phrases, the latter of which serves as critiques that can influence the recommendations through operations in the latent space. Others explore the interpretation of attributes \citep{Nema2021-ad, Balog2021-cj} to support critiquing. 
These works provide the basis for natural-language critiques as a vehicle for the user-feedback that REGEN provides.

\subsection{Recommendation Tasks}
\subsubsection{Sequential Item Recommendations}
There is a plethora of ongoing research in an effort to combine LLMs with recommender systems for sequential recommendations \citep{lin2021m6, Li2024-cx, sidahmed2022generating, zhang2021language, hebert2024flarefusinglanguagemodels, doddapaneni2024userembeddingmodelpersonalized, hybridbert4rec, Harte2023-gq, zhou2020s3, 10.1145/3523227.3546777, qiu2021u, zhuo2022tiger, raffel2020exploring, kddLLMCF, userllm}. These models explore different approaches for representing item and user interactions (e.g. ID embeddings, text embedddings), while some of them seek to enhance quality by incorporating item and/or semantic context. Item recommendation is an important, primary task that REGEN supports. As it is based on the Amazon Reviews dataset, a popular benchmarking dataset, we believe REGEN is a natural fit for extending existing LLM-based sequential recommenders to also generate language.

\subsubsection{Item Recommendation and Text Generation}
There are a number of recommendation models that output both item recommendations and natural language.
P5 \cite{p5rec} represents all user-item signals, along with associated context, as natural language sequences that are input to an encoder/decoder model. It proposes several task families for item prediction and text generation. User-LLM \citep{userllm} encodes user history into user embeddings, and incorporates them into an LLM via cross-attention. User-LLM supports three tasks, next item recommendation, genre prediction and reviews generation. While those projects propose tasks that involve both recommendations and text generation, they are separate tasks. In contrast, REGEN proposes integrating both those tasks, along with incorporating user feedback, into a single coherent task, while generating a rich and diverse set of narratives.

\section{REGEN}
This section details the REGEN dataset, including the generation process for the critiques and narratives, and evaluation methodology with a rater LLM. Further details of the modeling tasks for benchmarking are described in the following section.

\subsection{Amazon Reviews}
We use datasets from Amazon Product Reviews \citep{ni-etal} as a basis for augmentation with rich narratives. The Amazon Reviews dataset (2018) is a massive collection of customer reviews spanning various product categories. This version, released in 2018, contains over 233 million reviews, making it one of the largest publicly available datasets for sentiment analysis and recommendation systems. 
Each review includes item features like title, description, category, price and review features  including text, timestamp, score and summary. User sequences can be created by sorting each user review by timestamp, creating a sequential recommendation task.
For our work, we use ``Office Products'' and ``Clothing, Shoes and Jewelry,'' verticals. These are chosen to be representative samples with different item counts, Office with ~27k items and Clothing with ~376k items.
\begin{table}[!htb]
    \centering
    \caption{Statistics for the CALRec-processed Amazon Product Reviews dataset. I/U is the average number of items per user, or average sequence length.}
    \label{tab:dataset_stats}
    \begin{tabular}{l r r r}
        \toprule
        Dataset & \# Users & \# Items & I/U \\
        \cmidrule(r){2-4}
        Office & 89,489 & 27,887 & 7.86 \\
        Clothing & 1,168,735 & 376,847 & 9.21 \\
        \bottomrule
    \end{tabular}
\end{table}
\subsection{Dataset Generation}

\subsubsection{Data Preprocessing:}
User sequences $S_u = \{i_1, i_2, ..., i_n\}$ consisting of interactions $i$ are first created by aggregating reviews per user, and sorting them by timestamp. For comparable results, we use the CALRec~\cite{Li2024-cx} preprocessed datasets.
Notably, CALRec's de-duplication methodology helps mitigate trivial recommendation strategies, e.g. always recommending the most recent item. See Table~\ref{tab:dataset_stats} for the dataset statistics.

\subsubsection{Inpainting for Synthesizing Multi-Turn Conversations:}
The Amazon Reviews dataset, while valuable for recommendation studies, primarily reflects a series of individual user interactions rather than real conversations. To create a synthetic dataset that emulates multi-turn conversational dynamics, we employ an inpainting technique to generate the elements typically missing in non-conversational review data. Specifically, we leverage LLMs to synthesize these elements, using carefully designed prompts and an iterative evaluation process to ensure high-quality generation and minimize inaccuracies. This allows us to compare against existing research on these datasets for the recommendation task while adding natural language elements and setting new baselines for the language task.
In this work, we utilize the Gemini 1.5 Flash model, a fast and efficient version of the Gemini 1.5 model family~\cite{geminiteam2024gemini15unlockingmultimodal}.

We exploit the extended context length capabilities of the LLM to effectively process the complete user history $S_u$ for each user $u$. Our prompts are simple, employing a task prefix with instructions and the desired output format. This prefix is followed by the user's entire interaction history, including item metadata and review text. Finally, the prompt 
specifies the expected output format. We use the user history and associated reviews to generate critiques, purchase reasons and summaries. For product endorsement generation, however, we intentionally withhold the user's last review, enabling the model to learn how to endorse a recommended item without being influenced by that item's review.

\subsubsection{Critiques Generation}
REGEN adds generated critiques (short utterance) to steer the system from the current recommended item to a desired item. Our aim is to focus on the setting where a user refines from one item to a closely-related item variant (e.g., ``red ball-point pen'' to ``black ball-point pen''), rather than to another arbitrary item, which would be better served by other mechanisms, e.g. a new search query. Thus, REGEN only generates critiques between adjacent pairs of items that are sufficiently similar. We use the Amazon Reviews-provided hierarchical item categories as a rough proxy for item similarity and consider items sufficiently similar if at least four levels of the category match.

To generate critiques for adjacent item pairs that meet the similarity criteria, we query Gemini 1.5 Flash to synthesize several options, instructing the LLM to treat the first item as the current recommendation and the second item as representative of the user's desired goal. The prompt contains the item descriptions and few shot examples. We select at random one of the generated critiques to inpaint into the dataset for that pair. For pairs that do not meet the similarity criteria, REGEN contains a sentinel placeholder. Dataset users can decide to train on the placeholder, to model the case where end-users do not provide critiques emulating a new search query, or to replace them with other critiques.
For the ``Clothing'' dataset, REGEN includes LLM-generated critiques for about 18.6\% of adjacent items appearing in user sequences. For ``Office'' dataset there are about 17.6\%.

\subsubsection{Narrative Generation:}
We aim to generate diverse narratives for conversational recommender systems, varying in:
\begin{itemize}
    \item \textbf{Contextualization:} Narratives based on both recent and aggregate contextual information (e.g.,
user summaries vs explanations, endorsements of the most recent purchase).
    \item \textbf{Length:} Short-form and long-form narratives for varied conversational scenarios.
\end{itemize}
Using LLMs as automated evaluators, we assess generated critiques and narratives across multiple attributes. We demonstrate our approach using the Amazon product review dataset \cite{ni-etal} and evaluate with Gemini Pro. Detailed descriptions and examples of the narratives are summarized in Table~\ref{tab:nloutputs}.

\begin{table*}[!htb]
\caption{Generated narratives and their descriptions, with examples.}
\label{tab:nloutputs}
\centering
\footnotesize
\begin{tabular}{ m{1.7cm}|m{2.5cm}|m{4.0cm}|>{\em}m{7.5cm}<{\normalfont} } 
 \toprule
 \textbf{Context} & \textbf{Generated output} & \textbf{Description} & \textbf{Example} \\
 \midrule
 \multirow{10}{*}{Recent/Item} & Product Endorsements & \raggedright A tailored endorsement/sales pitch crafted based on purchase history and reviews & "This Michael Kors Bedford Top Zip Pocket Tote Bag is a luxurious and functional choice, perfect for someone who appreciates quality and style. Based on your previous purchases of  comfortable clothing and stylish sandals, this bag complements your taste for both everyday practicality and elegant accents. The durable leather, spacious interior with multiple pockets, and adjustable handles make it ideal for carrying your essentials in a sophisticated way.  It's a great investment in a timeless piece that will elevate your everyday looks and be a reliable companion for years to come."\\
 \cmidrule{2-4}
 & Purchase Reasons & \raggedright Concise explanations of the reasons behind a product recommendation. & "The customer wanted a stylish, relatively small tote bag with multiple pockets for organization." \\ 
 \cmidrule{2-4}
 & \raggedright Purchase Reason Explanations & Detailed and elaborate explanations of a purchase decision of the most recent item in context of the entire history. & "The customer had previously purchased a dress and was looking for a stylish handbag that would match. They mentioned the bag was smaller than expected, but was still stylish and had multiple pockets. This suggests the customer is looking for a functional and fashionable handbag." \\ 
 \midrule
 \multirow{8}{*}{Aggregate/User} & Brief User Summaries & \raggedright Concise summaries of a user's preferences and purchase history. & "This customer is a woman who enjoys shopping for both stylish and comfortable clothing. She is also interested in fashion accessories such as handbags." \\ 
 \cmidrule{2-4}
 & \raggedright Detailed User Summaries & \raggedright Comprehensive summaries of a user's preferences and history. & "This customer is a woman who has a taste for stylish and comfortable clothing. She has purchased sandals, dresses, and underwear, all items that suggest a preference for fashion and comfort. Her reviews also mention that she values quality, especially in clothing items like dresses. She is also interested in fashion accessories such as handbags. She seems to be an occasional buyer with a more practical and less frequent shopping approach." \\ 
 \cmidrule{2-4}
 & User Profiles & \raggedright A short phrase describing the type of user in product/shopping context. & "Fashionable and practical woman" \\
 \bottomrule
\end{tabular}
\end{table*}

\begin{table*}[!htb]
\caption{Rating attributes generated by the auto LLM rater to evaluate quality of critiques and narratives.}
\label{tab:combined_rating_dimensions}
\centering
\footnotesize
\begin{tabular}{ m{3.5cm}|m{3cm}|m{8cm} }
\toprule
\textbf{Category} & \textbf{Rating Attribute} & \textbf{Description} \\
\midrule
\multirow{3}{*}{Critiques} & Quality & Effectiveness of the refinement query in guiding the recommender towards desired Destination Item effectively. \\
\cmidrule{2-3}
& Specificity & Precision of the query in narrowing the search. \\
\cmidrule{2-3}
& Relevance & Meaningfulness and typicality of the query's attributes for finding the desired item. \\
\midrule
\multirow{4}{*}{Narratives} & Veracity & Accuracy and truthfulness of the purchase reason, analyzing specific evidence and inconsistencies within the user's purchase history. \\
\cmidrule{2-3}
& Personalization & Extent to which the narrative prioritizes information directly from the user's reviews and statements, as opposed to relying on product descriptions or general assumptions. \\
\cmidrule{2-3}
& Persuasiveness & LLM's ability to act as a skilled salesperson, crafting language that effectively promotes a product to the user. \\
\bottomrule
\end{tabular}
\end{table*}


\begin{figure*}
    \centering
    \begin{subfigure}{0.3\textwidth}
        \centering
        \includegraphics[width=\linewidth]{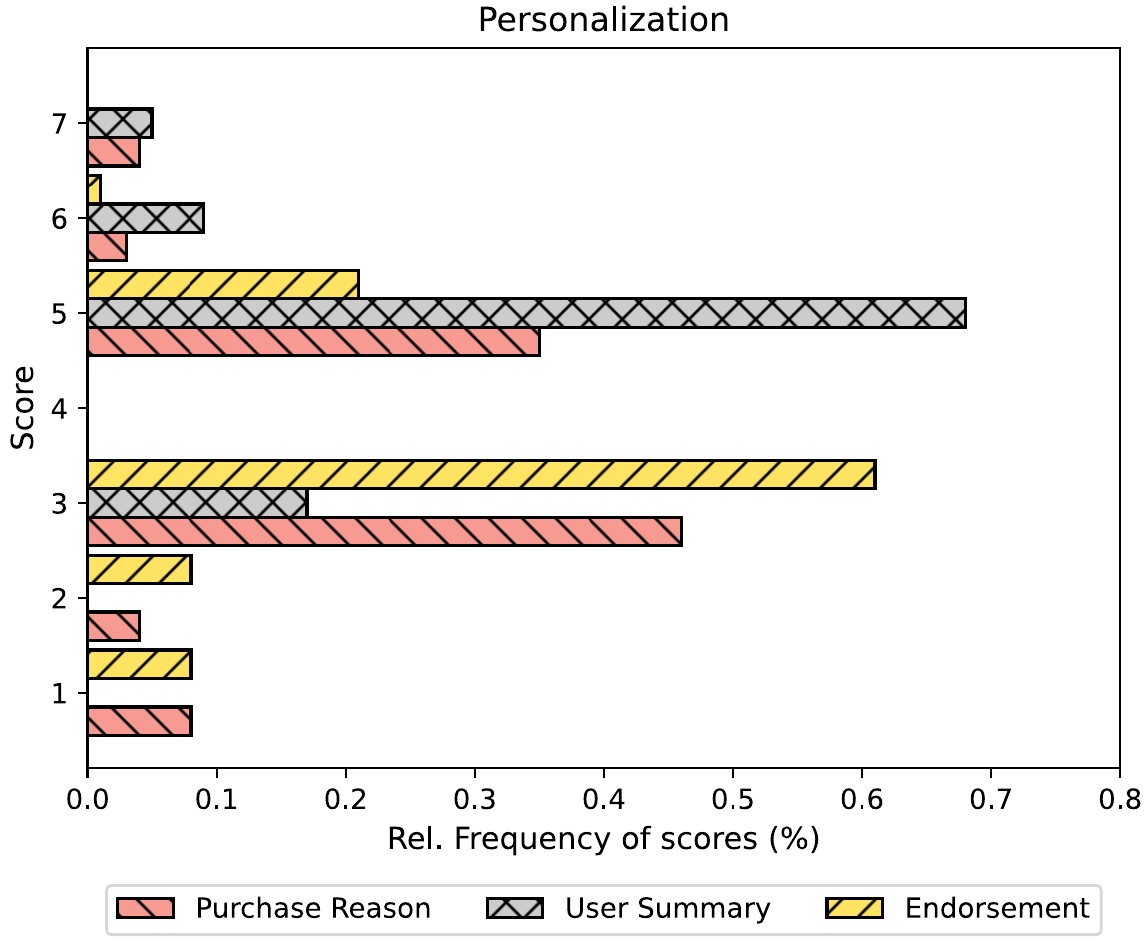}
        \label{fig:p13n}
    \end{subfigure}
    \begin{subfigure}{0.3\textwidth}
        \centering
        \includegraphics[width=\linewidth]{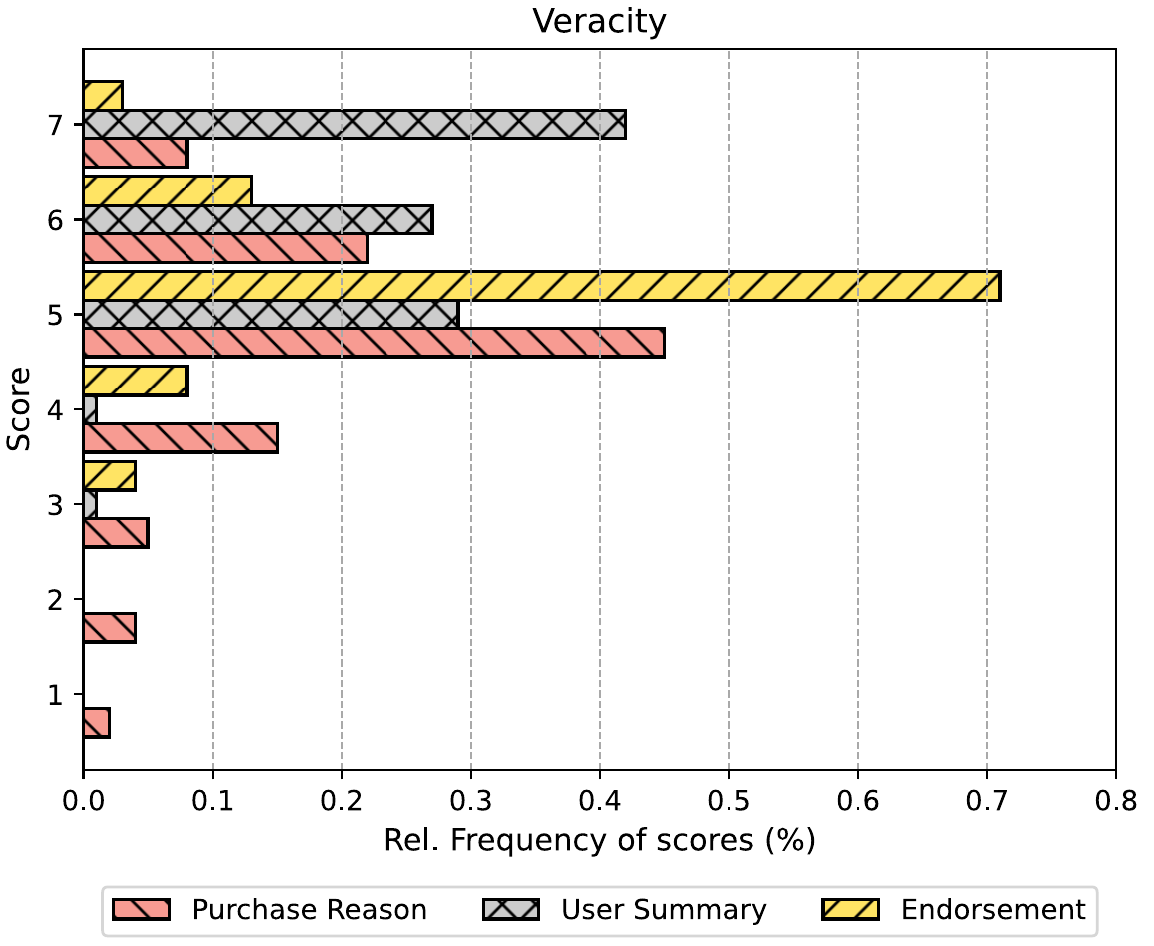}
        \label{fig:veracity}
    \end{subfigure}
    \begin{subfigure}{0.3\textwidth}
        \centering
        \includegraphics[width=\linewidth]{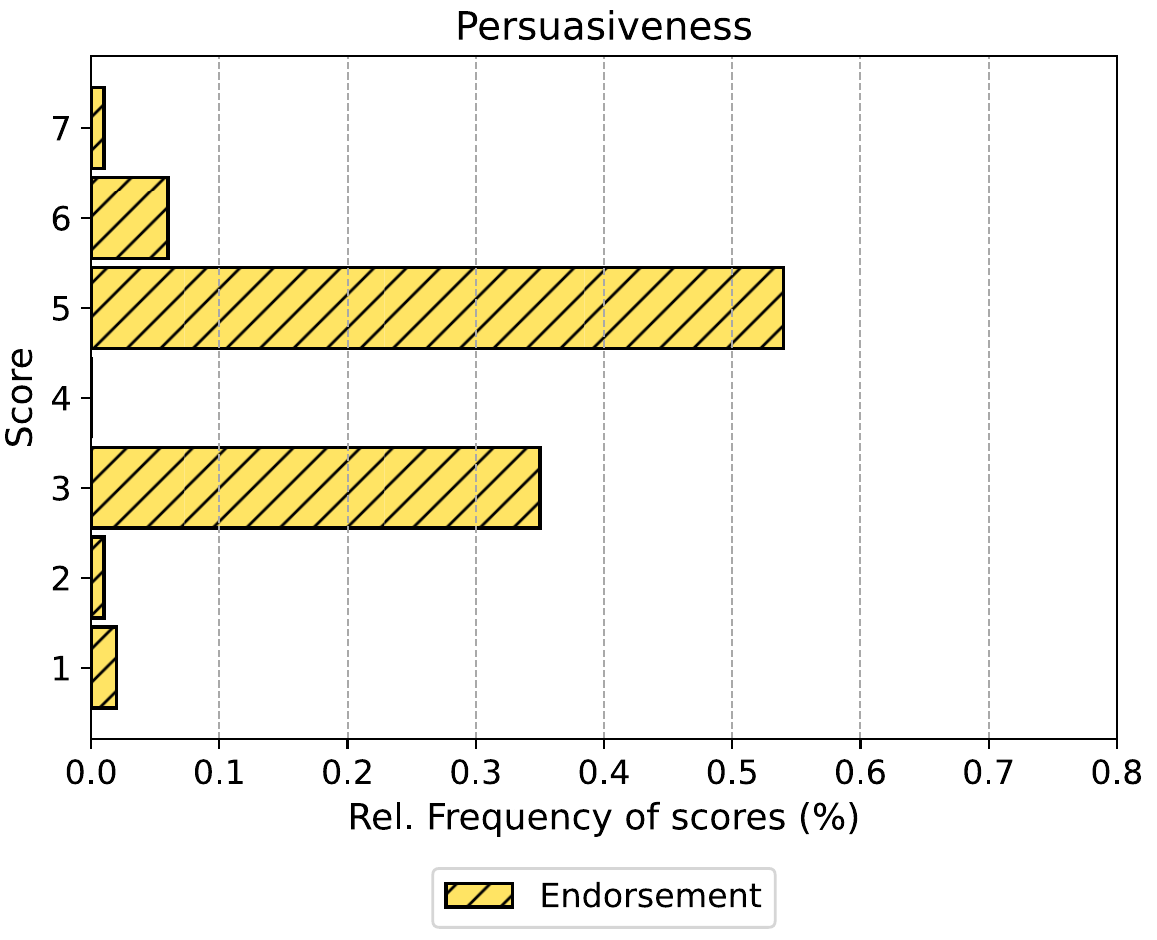}
    \end{subfigure}
    \caption{Narratives evaluated by rated attributes i) Personalization ii) Veracity and iii) Persuasiveness for Office Products}
    \label{fig:narratives_eval}
\end{figure*}

\begin{figure}
    \centering
    \includegraphics[width=0.8\linewidth]{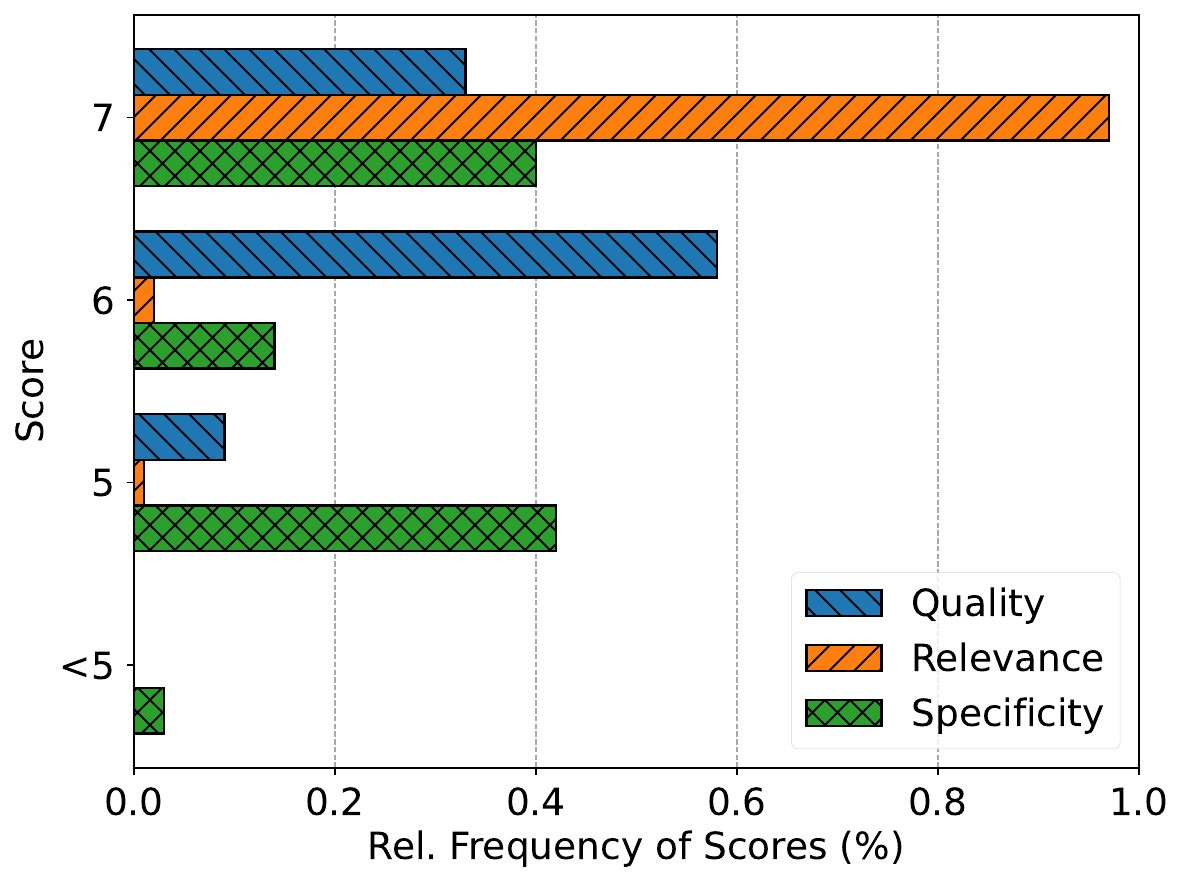}
    \caption{Critiques evaluated for i) Relevance ii) Specificity and iii) Quality for Office Products}
    \label{fig:steering_eval}
\end{figure}

\begin{figure}
    \centering
    \includegraphics[width=0.95\linewidth]{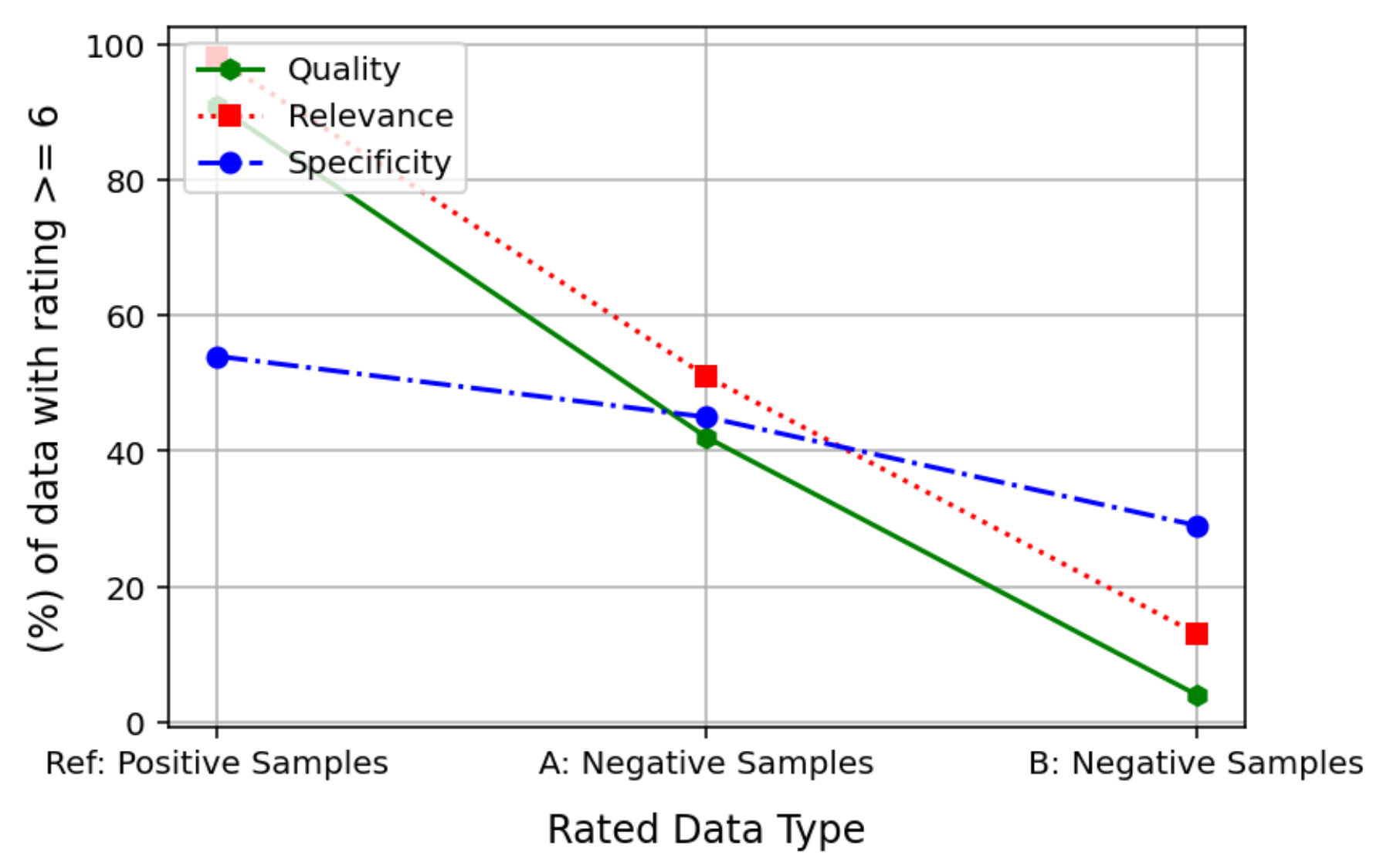}
    \caption{Steering Evaluation Bias with Negative Sampling on Office Products. Given a reference tuple (History, Critique, Next Item) A) Randomizes the History and B) randomizes the (History, Next Item) to a different user. This shows that scores are much lower on negatives and rules out positive bias in the auto rater.}
    \label{fig:steering_bias}
\end{figure}

\subsection {Evaluation using an Auto Rater}
\label{sec:eval}
We employed Gemini Pro LLM as an auto-rater. The rater received the task, conversation history, generated outputs, and was prompted to score outputs based on defined attributes. In this section, we show evaluation metrics and method for Office Products dataset. We have seen similar behaviors for the larger Clothing dataset overall, which are excluded due to space constraints.  

\subsubsection{Overall Approach}
To enhance reliability, we utilized an ensemble rating approach. Each output was rated seven times, and the scores were aggregated into a final score. We use a majority rule, where if over half of the ensemble runs agree on a score, that score becomes the final score. In the absence of a majority the average score (rounded down) is used.
While computationally demanding, this method yielded more robust attribute scores. Score distributions across all conversations are presented, and corresponding user scores are included in the released dataset. Furthermore, given that new attribute fields are in-painted across the entire user sequence, we captured score distributions by randomly selecting attribute values within each user's sequence and aggregating across all users. This approach provides a comprehensive view of the attribute score distribution over the generated sequences. The definitions of attributes are summarized in Table~\ref{tab:combined_rating_dimensions}.

\subsubsection{ Evaluation Results}
For narratives, we prompt the rater LLM to score on three attributes (veracity, personalization and persuasiveness) using a seven point likert scale. All narratives are evaluated for veracity and personalization, while product endorsement is also evaluated for persuasiveness. Further, we conduct the evaluation with three calls to the LLM, one per narrative group i) Purchase reason and explanation ii) User Summaries and iii) Product endorsement. This approach is used to avoid any correlation caused by Chain of Thought (CoT) bias across different narrative types. 

Figure~\ref{fig:narratives_eval} shows the rating distribution grouped by attributes.
Summaries show higher personalization scores, while the other contextual narratives show lower scores. This is expected as personalizing to a local item context is inherently more difficult. For veracity, which checks for hallucinations and grounding, scores are relatively higher across all narratives. 
Persuasiveness is evaluated for product endorsement and shows some distribution at lower scores ($\sim$35\% at score=3), which correlates to lower values of personalization for endorsement in some users, that can be attributed to a new product in the sequence. Overall, these results show that in this dataset, recommendations and narratives for certain users are influenced by popular behavior that their own history.

Critiques, formulated as short queries, were evaluated on three key attribute: i) Quality, ii) Specificity and iii) Relevance, measuring their effectiveness in guiding the recommender towards the next item, given the preceding conversation. Figure~\ref{fig:steering_eval} shows the distribution of scores, with higher than average scores across all the attributes.

\subsubsection{Evaluation Bias}
Figure~\ref{fig:steering_eval} shows high auto-rater scores across all attributes for critiques. To address potential positive bias, we generated negative samples: A) by swapping the history associated with a recommended item (resetting to a search scenario) and B) by swapping both the history and item (selecting them from a different user). Actual history, critique, item and corresponding negative samples are represented as  $(\tilde{H}, C, I)$ and $(\tilde{H}, C, \tilde{I})$ respectively.  Figure~\ref{fig:steering_bias} shows the ratio of high scores ($\geq 6$), which are lower for fully randomized samples (B) and slightly higher for history-swapped samples (A), as expected. Specificity demonstrates lower correlation to history and item, as it measures the precision of the critique itself to some extent.

\begin{figure*}[!htb]
    \centering
    \begin{subfigure}{0.5\textwidth}
        \includegraphics[width=\linewidth]{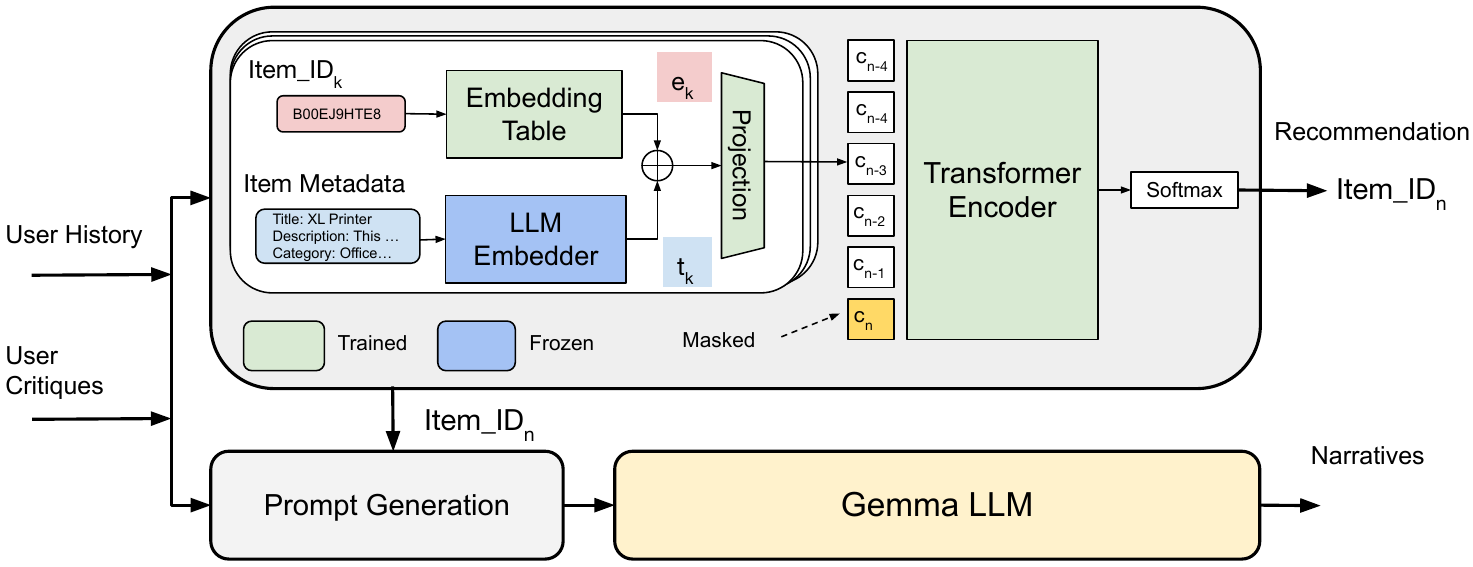}
        \caption{Hybrid Sequential Recommender (FLARE) + LLM}
        \label{fig:model_a}
    \end{subfigure}
    \hspace{5mm}
    \begin{subfigure}{0.4\textwidth}
        \includegraphics[width=\linewidth]{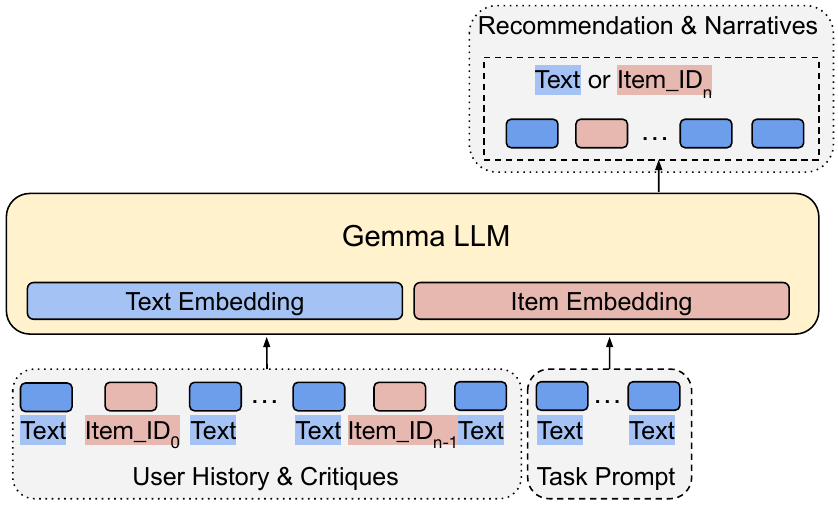}
        \caption{LUMEN: LLM for Critiquing and Generative Recommendation}
        \label{fig:model_b}
    \end{subfigure}
    \caption{Modelling Architectures used for Benchmarks}
    \label{fig:model}
\end{figure*}

\section{Benchmarking Methodology}
This section details the task definitions and modeling architecture.

\subsection{Task Definitions}
\label{sec:tasks}
We formalize the conversational recommendation task within the framework of user interaction sequences.  Consider a dataset of sequences $S = \{i_1, i_2,..., i_n\}$, where each interaction $(i \in \mathcal{I})$ is represented by the tuple $(\mathrm{ID}, T, P, C, \mathcal{T})$. Here, $\mathrm{ID}$ denotes the unique item identifier, $T = \{t^1, t^2,..., t^m\}$ represents a set of natural language components describing the item (including title, detailed product description, category, and other textual metadata), $P$ represents user-specific personalized text associated with the item (e.g., reviews, ratings, or other user-generated content), $C$ represents an explicit user's critiquing command influencing \emph{subsequent} item selection and $\mathcal{T}$ is the narrative corresponding to the interactions up to $n$, including the current item.
We define the following tasks:

\begin{enumerate}[align=right,itemindent=0em,labelsep=2pt,labelwidth=1em,leftmargin=*]
    \item \textbf{Recommendation Task (Next item prediction):} Predict the next item $i_{n+1}$ given the user's interaction history $S$. This is the standard sequential recommendation objective.
    
    \item \textbf{Recommendation Task with Critiquing:} Predict the next item $i_{n+1}$ given the user's history $S$ and user critique $C$.
    
    \item \textbf{Conversational Recommendation Task:} Generate a recommendation $i_{n+1}$ and natural language narrative $\mathcal{T}_{n+1}$, conditioned on the preceding interactions $S$ and critiques $C$. This task emphasizes capturing user preferences, critiques and generating recommendations along with contextually relevant narratives.
    
\end{enumerate}

This paper centers on the second and third tasks, as they align directly with the proposed dataset. The first task has been extensively studied, and we include it as a baseline.

\subsection{Model Architectures}

We benchmark two architectures, illustrated in Figure~\ref{fig:model}: a hybrid sequential recommender augmented with an LLM for generative outputs, and LUMEN, a purely LLM-based generative recommender. These are described in detail below.

\subsubsection{Hybrid Sequential Recommender + LLM:}
This approach integrates a traditional sequential recommender with an LLM, depicted in Figure~\ref{fig:model_a}. The sequential recommender focuses on modeling the collaborative filtering signal (encoded in the user-item interaction history), while the LLM is responsible for deeper semantic understanding and narrative generation. The sequential recommender produces a recommendation given user history, including content metadata for each item. We use FLARE~\cite{hebert2024flarefusinglanguagemodels} as the hybrid sequential recommender, which represents each item using both a learned embedding and a content-based encoding of the metadata (e.g. item title, description) using a language model. A perceiver resampler~\cite{perceiver} combines the two to yield a single embedding for each item. This architecture allows steering with natural language within a sequential recommendation task. The sequence is passed through a standard transformer stack, trained using standard masked-language-model (e.g. BERT4Rec~\cite{bert4rec}). The output recommendation is both shown to the user and conditions the LLM component. The LLM generates narratives, conditioned also on the user's history. This architecture emulates systems that integrate conventional recommenders with LLMs, which can be trained together or seperately. In our experiments, we train these components separately. 
To generate narratives, we prompt the LLM with the user history, consisting of a sequence of interactions, and any user critiques. The following structure is used to represent each item in the user's purchase history:

\begin{tcolorbox}
\small
\textbf{Input Prompt:} This is the summary of a user’s purchase history. First item purchase is START ID: <id> Title: <title> Category: <category> Description: <description> Price: <price> Rating: <rating> User Review: <review> User Critique: <user critique> END... Next item purchase is START ... END ... The last item purchase is: <id (from recommender)> .... Predict the <narrative type> for the last purchased item.\\
\textbf{Target:} <narrative>
\end{tcolorbox}

\subsubsection{\textbf{LUMEN}: Critiquing, Generative Recommendation and Narratives with LLM:}
Figure~\ref{fig:model_b} illustrates LUMEN, a modeling framework where the LLM is trained to both comprehend critiques and function as a generative recommender, in addition to its narrative generation capabilities. This is achieved by augmenting an additional embedding table in the language model to enable direct generation of item ID tokens.  Specifically, the new embedding table, learned through softmax loss on item vocabulary, is utilized to predict these ID tokens.
Our approach draws inspiration from SemanticIDs \cite{NEURIPS2023_20dcab0f_SemanticID}, which represent items as token sequences derived from semantic embeddings and refined during LLM training. However, while SemanticIDs employs precomputed hierarchies in tokenization, LUMEN utilizes randomly initialized embeddings and single-token IDs, which are fully trained alongside the LLM. We find that this simplification works well given the vocabulary sizes of our benchmark datasets, and allows training a single end-to-end model which improves reproducibility of the benchmarks. 

To manage the generation of both item IDs and language tokens, we maintain separate vocabularies. During inference, the decoder dynamically selects the appropriate vocabulary, using a specific token to signal item ID generation.
The prompt structure is consistent with the previous approach, with the exception that a placeholder token (ID\_TOKEN) is used to denote the position of the item ID. This approach is analogous to the techniques employed for multimodal inputs and outputs in Gemma~\cite{gemmateam2024gemmaopenmodelsbased}. The input and output prompts used are structured as follows:

\begin{tcolorbox}
\small
\textbf{Input Prompt:} This is the summary of a user’s purchase history. First item purchase is START ID: ID\_TOKEN Title: <title> Category: <category> Description: <description> Price: <price> Rating: <rating> User Review: <review> User Critique: <user critique> END. Next item purchase is START ID\_TOKEN... END... The next item purchase, and the product endorsement are:

\textbf{Target:} ID: ID\_TOKEN  Text: <narrative>
\end{tcolorbox}

During training, the ID\_TOKEN placeholders are replaced with their corresponding item ID embeddings from the item vocabulary.  During decoding, we switch from text to item vocab to generate a ID\_TOKEN when a START token is detected. Once the ID\_TOKEN is generated, we switch back to the text vocabulary and then continue autoregressive decoding of text tokens to generate the associated narrative.
Our hypothesis is that this method enables the LLM to learn a meaningful item embedding space within the context of conversational recommendation, which is aligned with the semantic space represented by the text. We evaluate this hypothesis by comparing the performance of these two architectures across a series of benchmarks.

\begin{table*}[htbp]
\centering
\footnotesize
\caption{Office Products Dataset - Benchmarks with Joint Recommendation and Generation Tasks. To evaluate upper bound performance, we provide the true next item $\mathrm{ID}_{t+1}$ when generating Flare+LLM narratives.}
\begin{tabular}{l|l|l|l|l|l|l|l|l|l|l|l}
\toprule
\multirow{4}{*}{\shortstack{Conversational \\Task}} & & \multicolumn{2}{c}{Purchase} & \multicolumn{2}{c}{User} & \multicolumn{2}{c}{Product} & \multicolumn{2}{c}{Purchase} & \multicolumn{2}{c}{Long User} \\
 & & \multicolumn{2}{c}{Reason} & \multicolumn{2}{c}{Summary} & \multicolumn{2}{c}{Endorsement} & \multicolumn{2}{c}{Reason Expl.} & \multicolumn{2}{c}{Summary} \\ \cmidrule{3-12}
 & Metrics & Flare+LLM & LUMEN & Flare+LLM & LUMEN & Flare+LLM & LUMEN & Flare+LLM & LUMEN & Flare+LLM & LUMEN \\
\midrule
\multirow{6}{*}{\shortstack{Hist. \\ $\downarrow$ \\ID + Narrative}}
 & Recall@10 & 0.124	& 0.098 & 0.124 & 0.10 & 0.124 & 0.10 & 0.124	& 0.091	& 0.124 & 0.098 \\
 & NDCG@10 & 0.0976 & 0.078 & 0.0976 & 0.08 & 0.0976 & 0.08 & 0.0976 & 0.071 & 0.0976 & 0.078 \\
 & MRR & 0.0925 & 0.072 & 0.0925 & 0.074 & 0.0925 & 0.074 & 0.0925 & 0.064 & 0.0925 & 0.072 \\
\cmidrule{2-12}
 & BLEU & 19.99 & 5.43 & 15.05 & 12.47 & 16.77 & 5.34 & 20.2 & 10.34 & 15.14 & 11.4 \\
 & ROUGE & 41.3 & 17.3 & 34.9 & 31.8 & 31.2 & 19.9 & 40.2 & 27.5 & 31.7 & 26.4 \\
 & Sem. Sim. & 0.833 & 0.642 & 0.87 & 0.849 & 0.927 & 0.742 & 0.88 & 0.743 & 0.905 & 0.858 \\ 
\midrule
\multirow{6}{*}{\shortstack{Hist. + Critique  \\ $\downarrow$ \\ ID + Narrative}}
 & Recall@10 & 0.1402 & 0.123 & 0.1402 & 0.117 & 0.1402 & 0.124 & 0.1402 & 0.122 & 0.1402 & 0.122  \\
 & NDCG@10 & 0.113 & 0.102 & 0.113 & 0.098 & 0.113 & 0.103 & 0.113 & 0.1 & 0.113 & 0.1 \\
 & MRR  & 0.1076 & 0.095 & 0.1076 & 0.092 & 0.1076 & 0.097 & 0.1076 & 0.094 & 0.1076 & 0.094 \\
\cmidrule{2-12}
 & BLEU   & 19.0 & 6.21  & 15.46 & 12.52 & 16.19 & 5.61  & 20.15 & 11.09 & 15.89 & 11.98\\
 & ROUGE  & 41.3 & 19.0  & 36.5 & 31.9  & 30.7 & 20.0    & 40.0 & 28.8  & 33.3 & 27.8 \\
 & Sem. Sim. & 0.816 & 0.651 & 0.87 & 0.851 & 0.929 & 0.747 & 0.891 & 0.757 & 0.897 & 0.87
\\
\bottomrule
\end{tabular}
\label{tab:joint_task_results_office}
\end{table*}

\begin{table}[!ht]
\centering
\footnotesize
\caption{Clothing Dataset - Benchmarks with Joint Recommendation and Generation Tasks. To evaluate upper bound performance, we provide the true next item $\mathrm{ID}_{t+1}$ when generating Flare+LLM narratives.}
\begin{tabular}{c l|c|c|c|c}
\toprule
& & \multicolumn{2}{c}{Purchase Reason} & \multicolumn{2}{c}{User Summary} \\ \cmidrule{3-6}
Task & Metrics & Flare+LLM & LUMEN & Flare+LLM & LUMEN \\
\midrule
\multirow{6}{*}{\shortstack{Hist. \\ $\downarrow$ \\ID + Narrative}}
& Recall@10 & 0.126	& \underline{0.127}	& \underline{0.126}	& 0.118 \\
& NDCG@10 & \underline{0.117} &	0.111 &	\underline{0.117} & 0.100 \\
& MRR & \underline{0.115}	& 0.106	&  \underline{0.115} & 0.094 \\
\cmidrule{2-6}
& BLEU  & \underline{19.2}	 & 10.2	 & \underline{16.4}	 & 12.8 \\
& ROUGE  & \underline{39.8}	 & 29.7	 & \underline{33.9}	 & 29.9 \\
& Sem. Sim.  & \underline{0.89}	 & 0.78	 & \underline{0.92}	 & 0.90 \\ 
\midrule
\multirow{6}{*}{\shortstack{Hist. + Critique  \\ $\downarrow$ \\ ID + Narrative}}
& Recall@10  & 0.136	 & \underline{0.143}	 & 0.136	 & \underline{0.151} \\
& NDCG@10  & \underline{0.124}	 & 0.122	 & 0.124	 & \underline{0.134} \\
& MRR   & \underline{0.122}	 & 0.115	 & 0.122	 & \underline{0.128} \\
\cmidrule{2-6}
& BLEU    & \underline{19.2}	 & 10.5	 & \underline{16.3}	 & 13.3 \\
& ROUGE   & \underline{40.0}	 & 30.2	 & \underline{34.1}	 & 30.3 \\
& Sem. Sim.  & \underline{0.89}	 & 0.79	 & \underline{0.92}	 & 0.90 \\
\bottomrule
\end{tabular}
\label{tab:joint_task_results_clothing_subset}
\end{table}

\section{Experiments}
For benchmarking, we implemented two model architectures: i) the FLARE sequential recommender~\cite{hebert2024flarefusinglanguagemodels} for recommendation generation, coupled with the Gemma 2B~\cite{gemmateam2024gemmaopenmodelsbased} LLM for narrative generation conditioned on the recommended item; and ii) LUMEN, utilizing Gemma 2B as the LLM, for joint recommendation and narrative generation. We evaluated these models on the REGEN dataset, focusing on the conversational tasks detailed in Section~\ref{sec:tasks}. Our evaluation spanned two categories: i) \emph{Office Products}, comprising $\sim$28k items, and ii) \emph{Clothing, Shoes, \& Jewelry}, comprising $\sim$376k items. This allowed us to assess model performance across datasets with significantly different item vocabulary sizes.

\subsection{Setup}

\subsubsection{Training and Evaluation}
We assessed recommendation performance using standard metrics: Recall@N, NDCG@N and MRR, with N set to 10. These metrics are widely adopted in recommendation tasks. For evaluating the quality of generated text, we employed BLEU and ROUGE to measure n-gram similarity. Additionally, we utilized a Semantic Similarity Metric, calculated as the cosine similarity between sentence embeddings generated by the Gecko embedding model~\cite{lee2024geckoversatiletextembeddings}. 
For partitioning the dataset into training, validation, and test sets, we adopted the leave-one-out approach~\cite{kang2018leaveoneouteval}, a common practice in sequential recommendation. Specifically, the first $N-2$ interactions were used for training, the first $N-1$ interactions with the last item held out were used for validation, and all $N$ interactions with the last item held out were used for testing.

\subsubsection{Implementation Details}
Our primary objective in benchmarking is to provide representative performance results using well-known architectures from recent research for conversational recommendation, rather than to advocate for a specific model configuration. While we added some variations that work well for the task and the dataset, and performed basic hyperparameter tuning within each framework, we acknowledge that more extensive optimization, including finer-grained tuning and the use of larger component models, could potentially yield improved results. We leave these for future investigation.

With the hybrid sequential recommender, we employed a Masked Language Modeling (MLM) task during training to enhance sample efficiency. We used the Adam optimizer with $\beta_1=0.9$ and $\beta_2=0.99$, and adopted the hyperparameter configurations detailed in~\cite{hebert2024flarefusinglanguagemodels}. Further, we made a simplifying assumption that the ground truth item is available to the LLM. Modeling the error in item recommendation would require co-training or evaluating the sequential encoder and decoder LLM models together, which is difficult to reproduce. On the other hand, with this assumption, we view the results with this approach as an upper bound for the joint recommendation and generation tasks. The approach is also easily reproducible, allowing for comparisons of other component models.

With LUMEN, utilizing the LLMs for recommendation, we trained on subsequences up to length $N-2$ for the joint recommendation and generation task. We use the Gemma 2B model with a batch size of 512 and a constant learning rate of 1e-5, training for up to 10,000 steps with early stopping.

\begin{figure*}[!htb]
    \centering
    \includegraphics[width=0.98\linewidth]{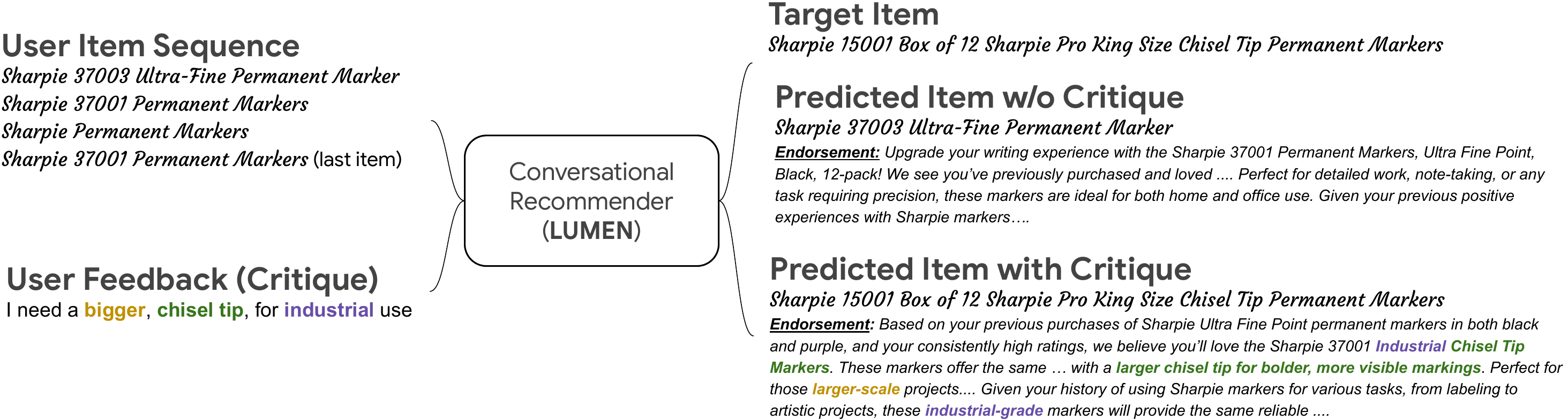}
    \caption{An example trace from LUMEN, using a joint generative task on REGEN, demonstrates the conversational recommender model's ability to process purchase interactions and in-context natural language critiques. This processing improves the predicted item and generates a natural language endorsement that aligns with the prediction.
    }
    \label{fig:lumen_trace}
\end{figure*}

\subsection{Results and Discussion}
Our experimental results, summarized in Table~\ref{tab:joint_task_results_office} and Table~\ref{tab:joint_task_results_clothing_subset}, lead to the following key findings:

\subsubsection{Recommendation Performance}
To establish baselines, we first compare recommendation metrics, which are well-studied for several recommender models in related work on amazon datasets. 

On REGEN Clothing (Table~\ref{tab:joint_task_results_clothing_subset}), compared to FLARE (See \cite{hebert2024flarefusinglanguagemodels} and references within) as a baseline, LUMEN trained on joint task achieves competitive performance despite generating recommendations and narratives within a single model, often achieving within $\pm 0.01$ Recall@10 and MRR, and improving performance in some cases. 

For Office Products (Table~\ref{tab:joint_task_results_office}) however, we see slightly lower recommendation metrics, within 10-15\% of the baseline. The Office dataset has a significantly smaller vocabulary ($\approx$27k) and training data size ($\approx$90k) (Table~\ref{tab:dataset_stats}). We believe this indicates that a sufficiently large training dataset is required to retain the recommendation performance in models trained on joint tasks.

\subsubsection{Recommendation Performance with Natural Language Critiques}
Incorporating user critiques consistently enhances recommendation performance, with improvements of +0.01--0.02 in Recall@10 and NDCG@10 and +0.01--0.03 in MRR observed for both hybrid and end-to-end LUMEN models ( Table~\ref{tab:joint_task_results_clothing_subset}).

\subsubsection{Narrative Performance}
As expected, LUMEN's narrative quality scores (BLEU, ROUGE) are lower than those of a hybrid model conditioned directly on ground-truth items. We attribute this performance gap primarily to recommendation errors, as LUMEN's narratives are based on its own predicted items in a single end to end model. We find that this dependency was less impactful for aggregate narratives such as ``User Summary," which rely less on the accuracy of the most recent item recommendation and more on the overall user history. In this case, LUMEN compares well with an LLM trained to generate narratives alone. Further, we expect that an integrated hybrid architecture where the output of the recommender is directly used as input to LLM is likely to see similar errors.

\subsubsection{Validation with Model Traces}
Figure~\ref{fig:lumen_trace} shows an example trace demonstrating recommendation, natural language understanding and generation capabilities of the LUMEN model trained on the joint tasks with the REGEN dataset.

\section{Limitations and Conclusion}
This paper introduced REGEN, a novel dataset enhancing Amazon Reviews with personalized narratives and critiques, facilitating a new task in conversational recommendation. While a comprehensive human evaluation of REGEN's critiques for ``naturalness'' would be beneficial, especially given their intended role as user feedback proxies, we argue that REGEN remains valuable even without it. Obtaining reliable human annotations for subjective, context-dependent textual tasks at scale is challenging. Moreover, the utility of a dataset is not solely dependent on perfect human alignment, as evidenced by the widespread use of Amazon Reviews despite its limitations. We demonstrate REGEN's value by showing its effectiveness in developing conversational recommendation models that integrate sequential recommendations, critique generation, and narrative generation. 

Our benchmarking efforts, while effective, utilized relatively simple architectures, including randomly initialized embeddings and single-token IDs. For larger item spaces, future work could explore optimizations like multi-token semantic IDs \cite{NEURIPS2023_20dcab0f_SemanticID} or other richer representations.

In summary, REGEN provides a dataset with consistent user preferences, recommendations, and generated narratives, enabling the study of LLM capabilities in conversational recommendation. We evaluated REGEN using LUMEN, an LLM-based model for joint recommendation and narrative generation, demonstrating its utility, along with sequential recommender models. We believe REGEN serves as a fundamental resource for studying the capabilities of conversational recommender models, a crucial step towards personalized multi-turn systems.

\bibliographystyle{ACM-Reference-Format}
\balance
\bibliography{main}

\appendix

\end{document}